\documentclass[letterpaper]{article} 
\usepackage[preprint]{aaai2027}
\usepackage[hyphens]{url}  
\usepackage{graphicx} 
\usepackage[numbers,sort&compress]{natbib}  
\usepackage{caption} 
\usepackage{booktabs} 
\usepackage{amsmath}
\usepackage{amssymb}
\usepackage{multirow}
\usepackage{tikz}
\usetikzlibrary{positioning}

\let\awxbibstyle\bibliographystyle
\renewcommand{\bibliographystyle}[1]{}

\title{Redefining the Pareto Frontier with AWX, a Swarm-Native\\
Agent Orchestration with Cost-Aware Multi-Tier Intelligence}

\newcommand{\awxauth}[4]{%
  \begin{tabular}[t]{@{}c@{}}
    #1\\[2pt]
    \normalfont\normalsize #2\\
    \normalfont\normalsize #3\\
    \normalfont\small\texttt{#4}
  \end{tabular}%
}
\author{
\awxauth{Akash Gupta}{Expedia Group}{Bengaluru, India}{akasgupta@expediagroup.com}%
\hspace{2.2em}%
\awxauth{Srinivas Billa}{Expedia Group}{London, UK}{nbilla@expediagroup.com}%
\hspace{2.2em}%
\awxauth{Xavi Amatriain}{Expedia Group}{San Jose, USA}{xavi@expediagroup.com}
}
\affiliations{}

\begin{document}
\let\bibliographystyle\awxbibstyle
\bibliographystyle{plainnat}
\setcitestyle{numbers,square,comma}
\setlength\bibhang{10pt}
\maketitle

\begin{abstract}
Production deployments of large language models face a cost-quality dilemma: frontier models are expensive at scale, while cheaper models fail unpredictably. We present Agent Weaver eXtreme (AWX), a domain-agnostic framework that coordinates heterogeneous LLMs through swarm-intelligence-inspired stigmergic signals. AWX attains accuracy comparable to state-of-the-art frontier LLMs at a fraction of their cost, redefining the Pareto frontier for multi-model systems.
\end{abstract}

\section{Introduction}

Large language models have achieved remarkable performance across vision, language, and reasoning tasks~\citep{openai2024gpt4v,anthropic2024claude3}, but production deployments expose a fundamental tension: single-model approaches suffer from unrecoverable systematic errors, while na\"ive multi-model ensembles achieve robustness at prohibitive cost.

Existing multi-agent frameworks~\citep{langchain2022,crewai2024,autogen2023,langgraph2024,metagpt2023} provide composable architectures for orchestration but lack three capabilities production deployments need. The first is \textit{cost-aware escalation} that routes easy items cheaply while reserving expensive computation for ambiguous cases. The second is \textit{principled diversity}, since agents receiving identical context produce redundant outputs~\citep{kuncheva2003measures}. The third is \textit{convergence signals} that determine when additional refinement yields diminishing returns.

Swarm intelligence~\citep{bonabeau1999swarm,dorigo1996ant} offers a principled solution: in ant colonies, simple agents coordinate complex foraging through \textbf{stigmergy}, indirect communication via environmental signals. We adapt these principles to multi-model orchestration, where quality indicators serve as digital pheromones that guide agent attention, signal convergence, and enable decentralized cost optimization.

The full AWX system composes twelve coordination mechanisms (Table~\ref{tab:mechanisms}), several adapting established methods (ACO and MAX-MIN, PSO with Clerc-Kennedy constriction, BDI agents) to language-model orchestration. The genuinely novel elements are digital stigmergy coordinating heterogeneous LLMs with no central controller, the deficit-guided weaving loop with Lyapunov-monitored stopping, and their integration into a cost-aware cascade. Its principal contributions are the following.

\begin{itemize}
\setlength\itemsep{2pt}
\item \textbf{Pheromone-guided stigmergic coordination} (Section~\ref{sec:pheromone}) with six signal types, MAX-MIN bounds~\citep{stutzle2000maxmin}, and Lyapunov-based convergence monitoring with a three-strike saturation rule.
\item \textbf{Progressive Signal Weaving} (Section~\ref{sec:weaving}), an iterative deficit-guided refinement in which particle swarm optimization~\citep{kennedy1995pso} explores an eight-dimensional strategy space following the residual-targeting principle of boosting~\citep{friedman2001greedy}.
\item \textbf{An Information-Asymmetric Agent Ensemble} (Section~\ref{sec:iaae}), in which Beliefs-Desires-Intentions agents~\citep{rao1995bdi} receive distinct pheromone-derived context, achieving diversity through information partitioning rather than model variation.
\item \textbf{Three-tier cascade escalation} (Section~\ref{sec:escalation}) in which 89.1\% of items resolve at Tier~1, at 4.3$\times$ lower cost with a 57.6\% disputed win rate over single-model baselines.
\item \textbf{Reproducible public-benchmark evidence} (Section~\ref{sec:publicbench}) on nine benchmarks, where task-adapted consensus gates place AWX on the cost-quality Pareto frontier in every case, exceeding the most accurate single-shot model on eight and tying on one at a fraction of frontier cost.
\item \textbf{Task-adapted agreement gates} (Section~\ref{sec:gatedefs}): a formal gate family spanning executable, behavioral, consensus, judge, and confidence signals, extended by \emph{corroboration ladders} on the escalated tail and \emph{kept-set disagreement checking}. The ladder and checker are new to LLM cascades, and the behavioral gate adapts execution-agreement selection~\citep{chen2023codet,li2022alphacode} into a cross-model routing signal.
\end{itemize}

\section{Related Work}
\label{sec:related}

\textbf{Multi-agent frameworks.} LangChain~\citep{langchain2022}, CrewAI~\citep{crewai2024}, AutoGen~\citep{autogen2023}, LangGraph~\citep{langgraph2024}, and MetaGPT~\citep{metagpt2023} provide composable abstractions for chaining model calls and modeling workflows as graphs. They excel at \textit{orchestration}, the static composition of agent steps, but not \textit{coordination}: none supplies a formal convergence criterion, a principled source of ensemble diversity, or cost-aware escalation. AWX treats coordination, rather than composition, as the primary design problem.

\textbf{Multi-model collaboration.} Mixture-of-Agents~\citep{wang2024moa} found that layered collaboration helps even with identical models, \citet{li2024more} that accuracy scales with agent count, \citet{du2024debate} that debate improves factuality, and self-consistency~\citep{wang2023selfconsistency} that majority vote over samples improves reasoning. On the cost axis, FrugalGPT~\citep{chen2023frugalgpt} cascades from cheap to expensive models, and GPTSwarm~\citep{zhuge2024gptswarm} optimizes agents as computational graphs. These methods typically assume a fixed topology or homogeneous sampling. AWX instead coordinates \textit{heterogeneous} models through shared stigmergic signals, without centralized orchestration logic.

\textbf{Learned orchestration.} Two recent systems learn the coordinator itself: the Conductor~\citep{nielsen2026conductor} trains a 7B reinforcement-learning orchestrator over worker LLMs, and \textsc{Trinity}~\citep{xu2026trinity} evolves a CMA-ES coordinator that assigns Thinker, Worker, and Verifier roles. Both, like AWX, exceed their strongest worker by coordinating heterogeneous models, but they train a task-general controller whereas AWX stays training-free, with stigmergic signals, a per-task gate search, and explicit cost playing the role of the learned policy.

\textbf{Swarm intelligence.} Stigmergy, indirect coordination through signals left in a shared environment, underlies Ant Colony Optimization~\citep{dorigo1996ant} and its MAX-MIN variant~\citep{stutzle2000maxmin}, while particle swarm optimization with the Clerc-Kennedy constriction factor~\citep{kennedy1995pso,clerc2002particle} provides formal convergence in continuous spaces. These methods have rarely been applied to coordinating language models. AWX adapts them to that discrete space, with bounded pheromone levels preventing premature convergence and constriction-factor PSO searching the strategy configuration space.

\textbf{Ensemble diversity.} Diverse, accurate base learners yield strong ensembles~\citep{dietterich2000ensemble}, yet no single diversity measure reliably predicts ensemble accuracy~\citep{kuncheva2003measures}. Most language-model ensembles induce diversity through model or prompt variation. AWX instead controls \textit{what each agent observes} through role-specific belief models, a mechanism orthogonal to model variation that the pheromone field makes explicit.

\section{Method}
\label{sec:method}

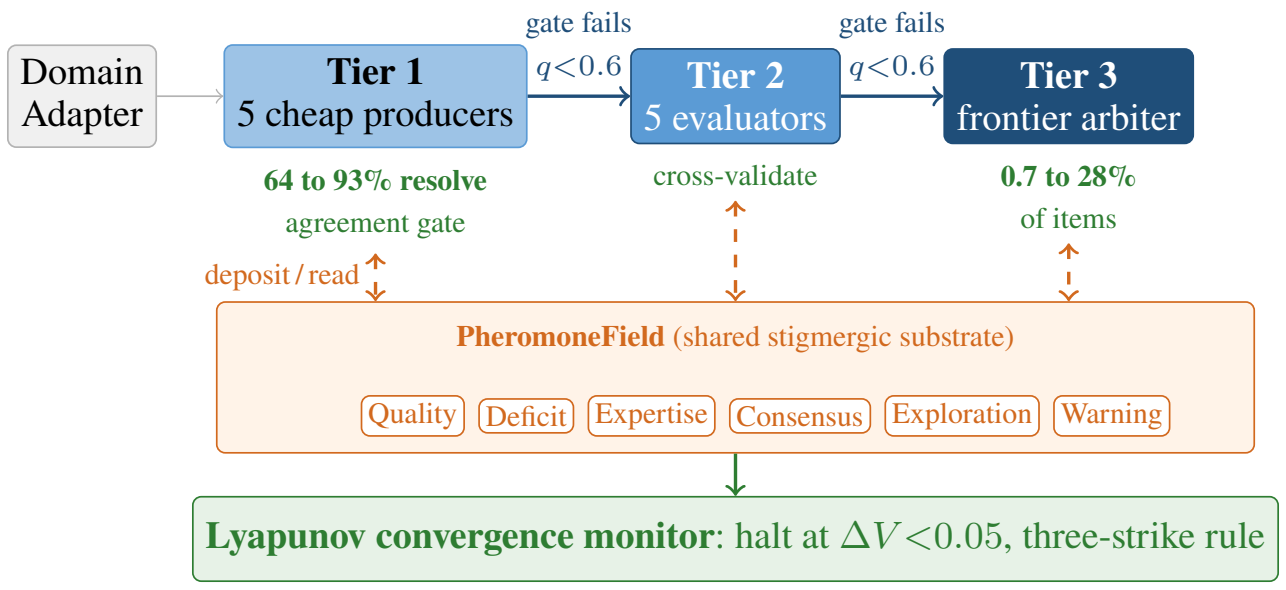
\begin{figure}[tb]
\centering
\definecolor{tierA}{HTML}{9DC3E6}\definecolor{tierB}{HTML}{5B9BD5}\definecolor{tierC}{HTML}{1F4E79}
\definecolor{awxor}{HTML}{D2691E}\definecolor{awxbg}{HTML}{FFF3E8}
\definecolor{okgr}{HTML}{2E7D32}\definecolor{okbg}{HTML}{E7F2E7}
\resizebox{\columnwidth}{!}{%
\begin{tikzpicture}[font=\footnotesize,
  box/.style={draw,rounded corners=2pt,minimum height=7mm,align=center,inner sep=3pt},
  chip/.style={draw=awxor,fill=white,rounded corners=2pt,inner sep=1.6pt,font=\scriptsize,text=awxor}]
\node[box,fill=gray!12,draw=gray!60] (da) {Domain\\Adapter};
\node[box,fill=tierA,draw=tierB,right=5.5mm of da] (t1) {\textbf{Tier 1}\\5 cheap producers};
\node[box,fill=tierB,draw=tierC,right=8.5mm of t1,text=white] (t2) {\textbf{Tier 2}\\5 evaluators};
\node[box,fill=tierC,draw=tierC,right=8.5mm of t2,text=white] (t3) {\textbf{Tier 3}\\frontier arbiter};
\draw[->,thick,tierC] (t1) -- node[above,font=\scriptsize,align=center]{gate fails\\$q{<}0.6$} (t2);
\draw[->,thick,tierC] (t2) -- node[above,font=\scriptsize,align=center]{gate fails\\$q{<}0.6$} (t3);
\draw[->,gray!60] (da) -- (t1);
\node[below=0.5mm of t1,font=\scriptsize,text=okgr,align=center] (r1) {\textbf{64 to 93\% resolve}\\agreement gate};
\node[below=0.5mm of t2,font=\scriptsize,text=okgr] (r2) {cross-validate};
\node[below=0.5mm of t3,font=\scriptsize,text=okgr,align=center] (r3) {\textbf{0.7 to 28\%}\\of items};
\node[box,fill=awxbg,draw=awxor,minimum width=86mm,minimum height=12.5mm,below=10mm of t2,yshift=-3mm] (pf) {};
\node[anchor=north,font=\scriptsize,yshift=-0.8mm] at (pf.north)
  {\textcolor{awxor}{\textbf{PheromoneField} (shared stigmergic substrate)}};
\node[chip,anchor=south west] at ([xshift=12mm,yshift=1.4mm]pf.south west) (c1) {Quality};
\node[chip,right=1.1mm of c1] (c2) {Deficit};
\node[chip,right=1.1mm of c2] (c3) {Expertise};
\node[chip,right=1.1mm of c3] (c4) {Consensus};
\node[chip,right=1.1mm of c4] (c5) {Exploration};
\node[chip,right=1.1mm of c5] (c6) {Warning};
\draw[<->,dashed,awxor,thick] (r1.south) -- node[left,font=\scriptsize,text=awxor]{deposit\,/\,read} (r1.south|-pf.north);
\draw[<->,dashed,awxor,thick] (r2.south) -- (r2.south|-pf.north);
\draw[<->,dashed,awxor,thick] (r3.south) -- (r3.south|-pf.north);
\node[box,fill=okbg,draw=okgr,minimum width=86mm,below=3.5mm of pf]
  (cv) {\textcolor{okgr}{\textbf{Lyapunov convergence monitor}: halt at $\Delta V{<}0.05$, three-strike rule}};
\draw[->,okgr,thick] (pf) -- (cv);
\end{tikzpicture}%
}
\caption{AWX architecture. Diverse producers, evaluators, and a frontier arbiter coordinate through the shared six-channel PheromoneField rather than a central controller. Only items whose agreement gate fails ($q{<}0.6$) escalate, so the arbiter sees 0.7 to 28\% of items, and a Lyapunov monitor halts refinement at saturation. A pluggable DomainAdapter keeps the engine domain-agnostic.}
\label{fig:arch}
\end{figure}

AWX coordinates heterogeneous agents through a seven-layer architecture in which the \textbf{PheromoneField} serves as the central coordination substrate, replacing centralized orchestration logic (Figure~\ref{fig:arch}). The framework is domain-agnostic: a pluggable \texttt{DomainAdapter} interface defines work-item schemas, quality dimensions, prompt templates, and evaluation rubrics for each target domain.

The complete system rests on twelve coordination mechanisms (Table~\ref{tab:mechanisms}). Three form the conceptual core the empirical results most directly probe: pheromone-guided coordination (Section~\ref{sec:pheromone}), progressive signal weaving (Section~\ref{sec:weaving}), and the information-asymmetric agent ensemble (Section~\ref{sec:iaae}), with three-tier cascade escalation (Section~\ref{sec:escalation}) binding them to cost. The emergence quotient is developed with the experiments.

\begin{table}[t]
\centering
\small
\begin{tabular}{@{}cl@{}}
\toprule
\textbf{\#} & \textbf{Coordination mechanism} \\
\midrule
1 & Pheromone-guided stigmergic coordination \\
2 & Information-asymmetric agent ensemble (BDI) \\
3 & Lyapunov-based convergence detection \\
4 & Particle-swarm strategy optimization \\
5 & Emergence detection via the SIQ \\
6 & Three-tier cost-aware cascade escalation \\
7 & Game-theoretic dynamic coalition formation \\
8 & Progressive signal weaving (deficit-guided) \\
9 & Domain-agnostic pluggable adapter orchestration \\
10 & Quorum-sensing consensus detection \\
11 & Intelligent dynamic agent-count estimation \\
12 & Swarm-scale provenance and traceability \\
\bottomrule
\end{tabular}
\caption{The twelve coordination mechanisms of AWX. Mechanisms 1, 2, and 8 with their dependencies 3, 4, and 6 form the conceptual core given formal treatment.}
\label{tab:mechanisms}
\end{table}

\subsection{Pheromone-Guided Stigmergic Coordination}
\label{sec:pheromone}

The core coordination mechanism is a shared \textbf{PheromoneField}, a data structure implementing \textit{digital stigmergy} through six typed signal channels per work item: \textbf{Quality} reinforces high-quality outputs, \textbf{Deficit} attracts refinement attention to quality gaps ($1.0 - q_i$), \textbf{Expertise} informs routing, \textbf{Consensus} signals evaluator agreement for quorum sensing, \textbf{Exploration} marks refined items to prevent redundant work, and \textbf{Warning} repels agents from error-prone outputs.

Pheromone dynamics follow modified Ant Colony Optimization rules with MAX-MIN bounds~\citep{stutzle2000maxmin}:
\begin{align}
\tau_d(t{+}1) &= (1 - \rho) \cdot \tau_d(t) + \textstyle\sum_k \Delta\tau_{d,k} \label{eq:update}\\
\Delta\tau_{d,k} &= \frac{\max(0,\; s_{\text{new}} - s_{\text{old}})}{\max(c_k,\; \epsilon)} \label{eq:delta}\\
\tau_{\min} &\leq \tau_d(t) \leq \tau_{\max} \label{eq:bounds}
\end{align}
where $\rho = 0.1$ is the evaporation rate (10\% per cycle), $s$ denotes quality scores, $c_k$ is agent~$k$'s cost in USD, $\epsilon = 0.001$, and the bounds $\tau_{\min} = 0.01$ and $\tau_{\max} = 5.0$ prevent signal extinction and premature convergence.

\subsubsection{Probabilistic Item Selection.}
Agent work selection follows the Ant Colony Optimization decision rule:
\begin{equation}
P(\text{agent}_i \text{ selects item}_j) = \frac{\tau_j^\alpha \cdot \eta_j^\beta}{\sum_k \tau_k^\alpha \cdot \eta_k^\beta}
\label{eq:selection}
\end{equation}
where $\alpha = 1.0$ is the pheromone influence, $\beta = 2.0$ is the heuristic influence, and $\eta_j$ is heuristic attractiveness (inverse confidence). The higher $\beta$ biases selection toward low-confidence items that need the most attention.

\subsubsection{Convergence Monitoring.}
We define a Lyapunov candidate function:
\begin{equation}
V(\mathbf{s}_t) = \frac{1}{N}\textstyle\sum_{i=1}^{N} \max(0,\; \theta_i - s_i(\mathbf{s}_t))
\label{eq:lyapunov}
\end{equation}
where $\theta_i$ is the quality target for entry $i$ and $N$ is the number of score entries (items times dimensions). Because each round can only raise or hold per-item quality, $V$ is non-increasing in practice, and we use it as an empirical convergence signal rather than a fixed-point guarantee. Convergence is declared when $V \leq 0.05$ or three consecutive rounds show $\Delta V < 0.05$ (the three-strike rule), with a minimum of two rounds enforced.

\subsection{Progressive Signal Weaving}
\label{sec:weaving}

Each weaving round executes a \textbf{strategy}, a coordinated sequence of agent actions defining the \textit{exploration-exploitation balance}, drawn from six options: \textbf{Progressive} (produce, evaluate, refine), \textbf{Consensus} (quorum sensing), \textbf{Polyphonic} (multi-voice synthesis), \textbf{Debate} (adversarial refinement), \textbf{Cascade} (hierarchical escalation), and \textbf{Hybrid}.

\subsubsection{Particle-Swarm Strategy Selection.}
An \textbf{EvolutionarySelector} uses particle swarm optimization~\citep{kennedy1995pso} to search an eight-dimensional strategy configuration space (strategy identifier, temperature, blend factor, deficit threshold, diversity and cost weights, agent count, maximum rounds), using the Clerc-Kennedy constriction factor~\citep{clerc2002particle}:
\begin{align}
v_i(t{+}1) &= \chi \left[ v_i(t) + c_1 r_1 (p_i - x_i) + c_2 r_2 (l_i - x_i) \right] \label{eq:pso_v}\\
x_i(t{+}1) &= x_i(t) + v_i(t{+}1) \label{eq:pso_x}\\
\chi &= \frac{2}{|2 - \phi - \sqrt{\phi^2 - 4\phi}|},\quad \phi = c_1 + c_2 \label{eq:chi}
\end{align}
where convergence requires $\phi > 4$. We use 20 particles with $c_1 = c_2 = 2.05$ (so $\phi = 4.1$), a ring topology for the local best $l_i$, and two heuristic cold-start rounds. The fitness function is
\begin{equation}
f = \frac{q \cdot r}{\max(c,\; 0.001)}
\label{eq:fitness}
\end{equation}
where $q$ is quality, $r$ is reliability (the fraction of successful agent calls), and $c$ is cost in USD.

\subsubsection{Deficit-Targeted Refinement.}
Within the Progressive strategy each round has three phases. \textbf{Produce}: agents read Deficit pheromones via probabilistic selection (Equation~\ref{eq:selection}) to target underrepresented categories. \textbf{Evaluate}: batch evaluation deposits a Quality pheromone $q_i$ and a Deficit pheromone $1.0-q_i$ per item. \textbf{Refine}: items below the deficit threshold ($\theta=0.7$) are refined concurrently, while items above 0.90 confidence bypass refinement. From the second round on, gap-aware re-production injects category-gap context into the producers.

\subsubsection{Quorum-Sensing Consensus.}
\label{sec:quorum}
The Consensus strategy formalizes agreement detection as \textit{quorum sensing}, borrowing from bacterial signaling: for a work item $j$ scored independently by $m$ agents as $\{s_{j,1}, \dots, s_{j,m}\}$, AWX computes the dispersion
\begin{equation}
\sigma_j = \sqrt{\tfrac{1}{m}\textstyle\sum_{k=1}^{m}\left(s_{j,k} - \bar{s}_j\right)^2}
\label{eq:quorum}
\end{equation}
and declares consensus when $\sigma_j \leq \sigma_q$ and a quorum fraction $\rho_q$ of the agents fall within one dispersion band of the mean ($\sigma_q = 0.15$, $\rho_q = 0.6$). Items that fail quorum route to the cascade (Section~\ref{sec:escalation}). Quorum sensing turns agreement into a thresholded signal, so a few confident outliers cannot manufacture false consensus.

\subsection{Information-Asymmetric Agent Ensemble}
\label{sec:iaae}

The ensemble manufactures \textit{diversity} through information partitioning rather than model variation. Each agent follows a Beliefs-Desires-Intentions architecture~\citep{rao1995bdi}, and the key asymmetry is \textit{role-specific pheromone access}: Producers read Deficit pheromones to find missing content, Evaluators observe raw quality across all dimensions, Refiners target items below the 0.7 threshold, and Synthesizers read Consensus pheromones to consolidate. Even on the same underlying model, agents produce diverse outputs because they observe different slices of the shared state, analogous to feature subsampling in random forests~\citep{breiman2001random} but on task context. AWX also draws agents from different providers to leverage distinct training data and failure modes.

\subsubsection{Game-Theoretic Dynamic Coalition Formation.}
Agents interact through a dynamic graph whose edge weights update via an exponential moving average:
\begin{equation}
w_{\text{new}} = 0.3 \cdot q_{\text{interaction}} + 0.7 \cdot w_{\text{old}}
\label{eq:ema}
\end{equation}
clamped to $[0, 1]$, where $q_{\text{interaction}}$ is the realized quality of a joint action. Coalitions form by a cooperative-game rule rather than central assignment: for a candidate coalition $C \subseteq A$ we score
\begin{equation}
v(C) = \mathrm{cov}(C)\,\big(1 + \overline{w}(C)\big),
\label{eq:coalition}
\end{equation}
where $\mathrm{cov}(C) \in [0,1]$ is the fraction of the swarm's distinct capabilities represented in $C$ and $\overline{w}(C)$ is the mean pairwise interaction weight, rewarding complementary skills and successful past collaboration. Coalitions merge only when their union is super-additive, and stable coalitions are recovered as connected components whose internal edges exceed 0.6, a tractable surrogate for the core of the cooperative game.

\subsection{Three-Tier Cascade Escalation}
\label{sec:escalation}

The cascade partitions available agents across at most three tiers by role. At \textbf{Tier~1} producers independently process items, and those whose agreement gate passes (aggregated quality $\geq 0.6$) are resolved with a Quality deposit. At \textbf{Tier~2} evaluators cross-validate unresolved items, with Deficit pheromones proportional to $(0.6 - q_i)$ guiding attention. At \textbf{Tier~3} a single frontier arbiter resolves the rest with full prior context. Because diverse models that independently agree exit at Tier~1, only genuinely ambiguous cases escalate, and the logic applies unchanged across domains.

\subsection{Domain Adaptation Layer}
\label{sec:domain}

AWX separates coordination logic from domain semantics through a \texttt{DomainAdapter} abstraction specifying a typed \texttt{WorkItem} schema, weighted quality dimensions, per-role prompt templates, and an evaluation rubric, so the same engine operates across vision, language, extraction, and generative tasks unmodified.

\subsection{Swarm-Scale Provenance and Traceability}
\label{sec:provenance}

Because coordination is decentralized, AWX records every event (agent actions, pheromone deposits, evaluations, escalations, decisions) as a node in an append-only DAG whose edges record causal enablement, so the full causal chain behind any output is the ancestor set of its decision node, recoverable in linear time. Provenance composes across federated swarms by reference.

\subsection{Task-Adapted Agreement Gates}
\label{sec:gatedefs}

The tier-1 \emph{agreement gate} is the mechanism our experiments probe most directly. For an item $x$ with producer outputs $a_1,\dots,a_m$, a gate $G_{\sigma,\theta}$ keeps $x$ (serving the consensus answer without escalation) iff an agreement signal $\sigma(a_1,\dots,a_m; x) \geq \theta$. The signal is chosen per task from a hierarchy of error-correlated observables: \emph{executable tests}, \emph{behavioral agreement} (programs kept when their outputs match on probe inputs), \emph{label or numeric consensus}, \emph{judge-score routing}, and \emph{self-confidence}. Score-valued tasks use \emph{quorum sensing} (bounded dispersion with a quorum near the mean), so confident outliers cannot manufacture consensus. Two constructions extend the gate beyond tier~1: the \emph{corroboration ladder} adopts an escalated answer only when a tier-1 producer or second strong model matches it, and the \emph{disagreement checker} re-answers a pre-flagged slice of the kept set to re-route disagreements upward. The ladder and checker are new to LLM cascades, and the behavioral gate adapts execution-agreement selection to cross-model routing (Section~\ref{sec:publicbench} measures each).

\section{Experimental Setup}

\subsection{Public Benchmark Protocol}
\label{sec:benchsetup}

We evaluate AWX on nine widely used public benchmarks spanning knowledge, mathematical reasoning, code generation, dialogue, instruction following, and vision. Table~\ref{tab:benchsetup} gives each dataset, metric, and the consensus gate the swarm instantiates on it.

\begin{table}[t]
\centering
\footnotesize
\setlength{\tabcolsep}{3.5pt}
\begin{tabular}{@{}lrlll@{}}
\toprule
\textbf{Benchmark} & \textbf{Items} & \textbf{Metric} & \textbf{Gate signal} \\
\midrule
MMLU & 14,034 & Accuracy & letter consensus (4) \\
GPQA Diamond & 198 & Accuracy & self-confidence \\
GSM8K & 1,319 & Exact match & numeric consensus (3) \\
MATH & 5,000 & Exact match & self-confidence \\
HumanEval & 164 & Pass@1 & behavioral agreement \\
LiveCodeBench & 976 & Pass@1 & public-test execution \\
CoQA & 7,983 & Token F1 & judge selection \\
MT-Bench & 80 & Judge score & judge-score routing \\
ImageNet & 50,000 & Top-1 acc. & label agreement (2) \\
\bottomrule
\end{tabular}
\caption{Public benchmark suite: MMLU~\citep{hendrycks2021mmlu}, GPQA Diamond~\citep{rein2023gpqa}, GSM8K~\citep{cobbe2021gsm8k}, MATH~\citep{hendrycks2021math}, HumanEval~\citep{chen2021humaneval}, LiveCodeBench~\citep{jain2024livecodebench}, CoQA~\citep{reddy2019coqa}, MT-Bench~\citep{zheng2023judging}, and ImageNet~\citep{deng2009imagenet}, each on its full evaluation set. The gate column names the tier-1 agreement signal (parenthesized voter counts).}
\label{tab:benchsetup}
\end{table}

\subsubsection{AWX Configuration.}
Every benchmark uses the cascade strategy of Section~\ref{sec:escalation}: cheap producers whose \emph{agreement gate} resolves items the swarm agrees on, with disagreements escalating through stronger models under the corroboration ladder and MMLU disagreement checker of Section~\ref{sec:gatedefs}. Each task's gate is the strongest error signal it affords (Table~\ref{tab:benchsetup}). Producer rosters draw on Gemma 4 31B, Kimi K2.5, Llama 4 Maverick, Nova Micro/Pro, DeepSeek V3.2, and Sonnet 4.5, and escalation tiers on GLM 5.2, GPT-5.5, and Claude Opus 4.5 to 4.8, with gates and thresholds set by the strategy search of Section~\ref{sec:weaving} over cached full-set runs. All models run at temperature 0.0 as single deterministic full-set runs, gates never observe ground truth, and truncated runs are excluded rather than imputed.

\subsubsection{Baselines.}
For every benchmark we run a panel of 8 to 18 single-shot comparators under the identical harness, prompts, and scorer, spanning Nova Micro up to frontier models (GPT-5.5, Claude Opus 4.7/4.8, GLM 5.2, Gemma 4, Kimi K2.5, DeepSeek, Llama, and Pixtral Large). The comparison target is the \emph{most accurate} single-shot model on each benchmark, which is task-dependent (GPT-5.5 or Opus 4.8 on three tasks each, GLM 5.2 reasoning on two, Gemma 4 on one), so no fixed baseline is privileged. All costs are re-priced from logged token counts on one canonical rate card.

\subsection{Production-Scale Evaluation: Image Classification and Ad-Relevance}

\textbf{Image classification.} We classify 11{,}072 images from a large consumer platform for marketing suitability (40 scene categories, 21 compliance checks, a marketability verdict). Five diverse Tier~1 producers double as Tier~2 evaluators, with Claude Opus 4.6 as arbiter and single-shot baseline under identical instructions. Ground truth is an independent 11-model Agent-as-a-Judge panel~\citep{zheng2023judging} over 7{,}719 images, with predictions anonymized.

\textbf{Ad-relevance.} We validate the same recipe on a production advertising workload: annotating booked-ad relevance for search queries from a large commercial search marketplace, scored as booked-relevance accuracy (BRS) against the booking event itself and penalizing unjudged ads. The label-consensus gate has Nova 2 Lite and Gemma 3 4B produce, Gemma 3 27B evaluate, and disagreements route to a Claude Opus 4.7 arbiter. Three runs pool 11{,}000 searches into 6{,}071 booked-ad judgments.
\label{sec:adrelsetup}

\subsection{Cross-Domain Deployment}

To test domain-agnostic coordination, we deploy AWX across 15 production domains (spanning recommendation, planning, listing-quality analysis, personalization, safety evaluation, code generation, and clinical extraction), each supplying only its schema, quality dimensions, prompts, and rubric through the \texttt{DomainAdapter}. The particle-swarm selector converges to different dominant strategies per domain, with agent counts estimated per task and clamped to 4 to 500.

\section{Results}

\subsection{Public Benchmark Evaluation}
\label{sec:publicbench}

\begin{figure*}[t]
\centering
\includegraphics[width=0.70\textwidth]{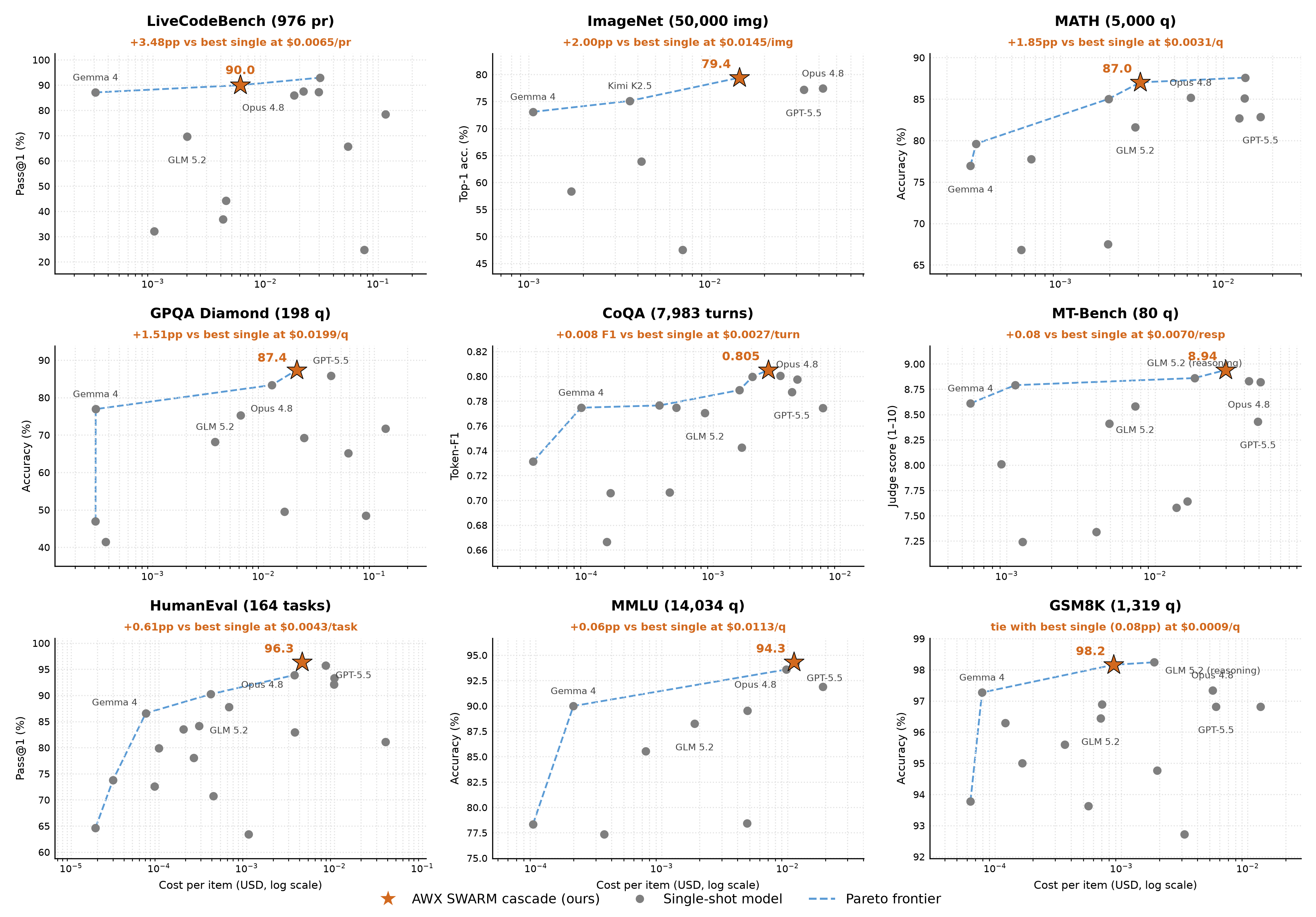}
\caption{AWX cascades vs.\ single-shot models: accuracy or quality versus cost per item (log scale) on all nine public benchmarks. Grey circles are single-shot models, the dashed line the Pareto frontier, and the orange star AWX, which tops or ties the frontier in every panel (cheaper to the left, better up). Each panel subtitle reports AWX versus the most accurate single-shot model on that benchmark.}
\label{fig:combinedpareto}
\end{figure*}

Table~\ref{tab:publicmain} and Figure~\ref{fig:combinedpareto} report the headline result. On all nine benchmarks AWX sits on the cost-quality Pareto frontier, and it exceeds the most accurate single-shot model on eight of the nine, tying within one standard error on GSM8K, at 14\% to 97\% of the strongest frontier model's per-item cost (median 34\%). Four margins exceed one binomial standard error (MATH, LiveCodeBench, ImageNet, CoQA F1). The rest (MMLU, HumanEval, GPQA at $n{=}198$, MT-Bench) are accuracy ties with a decisive cost advantage. Crucially, the best single model differs across tasks (GPT-5.5, Claude Opus 4.8, GLM 5.2 reasoning, or Gemma 4) while one coordination recipe tops all of them, so the gain is a property of coordination rather than of any constituent.

\begin{table}[t]
\centering
\footnotesize
\setlength{\tabcolsep}{3.2pt}
\begin{tabular}{@{}lrrlrr@{}}
\toprule
\textbf{Benchmark} & \textbf{AWX} & \textbf{\$/item} & \textbf{Best single} & \textbf{Score} & \textbf{\$/item} \\
\midrule
MMLU          & \textbf{94.30} & .0113 & GPT-5.5    & 94.24 & .0116 \\
GPQA Diamond  & \textbf{87.37} & .0199 & GPT-5.5    & 85.86 & .0407 \\
GSM8K         & 98.16 & .0009 & GLM 5.2 R  & \textbf{98.24} & .0018 \\
MATH          & \textbf{87.02} & .0031 & Opus 4.8   & 85.17 & .0063 \\
HumanEval     & \textbf{96.34} & .0043 & GPT-5.5    & 95.73 & .0080 \\
LiveCodeBench & \textbf{90.37} & .0065 & Gemma 4    & 86.68 & .0003 \\
CoQA (F1)     & \textbf{.8051} & .0027 & Opus 4.8   & .7975 & .0046 \\
MT-Bench      & \textbf{8.94}  & .0070 & GLM 5.2 R  & 8.86  & .0185 \\
ImageNet      & \textbf{79.41} & .0145 & Opus 4.8   & 77.41 & .0421 \\
\bottomrule
\end{tabular}
\caption{AWX versus the most accurate single-shot model per benchmark (8 to 18 comparators, identical harness, one rate card). AWX leads on eight of nine and ties on GSM8K within one standard error. GLM 5.2 R is the extended-reasoning configuration.}
\label{tab:publicmain}
\end{table}

\subsubsection{Gate Design and the Dominance Condition.}
\label{sec:gates}
The nine campaigns support an empirical design condition: the swarm dominates every single model when its agreement gate keeps a subset on which the consensus answer is at least as accurate as the best single model on that same subset. Gate signals form a quality hierarchy (Table~\ref{tab:gates}): executable and behavioral agreement yield the most precise kept sets and let consensus win the kept set outright, label or numeric consensus among decorrelated producers follows, and judge-score routing and self-confidence sit at the bottom. Two refinements convert near-misses into wins: the \emph{corroboration ladder} (100\% precise on the 45 HumanEval items it accepts) and the MMLU \emph{disagreement checker}, where on checker-flagged kept items the consensus is only 40 to 47\% correct against 63 to 68\% for the strong tier, so re-routing exactly those items flips the benchmark. GPQA, whose self-confidence gate sits at the bottom of the hierarchy, instead wins on the escalated arm. A negative result sharpens the picture: adding weak voters \emph{degrades} the gate because they correlate on wrong answers, and two strong-but-cheap decorrelated voters beat a crowd of six weaker ones.

\begin{table}[tb]
\centering
\footnotesize
\setlength{\tabcolsep}{3pt}
\begin{tabular}{@{}llrr@{}}
\toprule
\textbf{Gate signal} & \textbf{Benchmark} & \textbf{Kept prec.} & \textbf{Cons.$-$best} \\
\midrule
Executable tests     & LiveCodeBench & 95.6\% & $+1.2$ \\
Behavioral agreement & HumanEval     & 99.1\% & tie \\
Label/numeric        & GSM8K         & 98.9\% & $-1$ item \\
Label/numeric        & ImageNet      & 85.5\% & $+1.4$ \\
Judge-score routing  & MT-Bench      & n/a    & n/a \\
Self-confidence      & GPQA, MATH    & n/a    & n/a \\
\bottomrule
\end{tabular}
\caption{Gate-signal quality hierarchy (high to low). Kept-set precision and the margin of consensus over the best single model both fall with signal strength, so top signals win the kept set outright while self-confidence gates (GPQA, MATH) instead win on the escalated arm.}
\label{tab:gates}
\end{table}

\subsubsection{Escalation Efficiency.}
\begin{figure}[tb]
\centering
\includegraphics[width=\columnwidth]{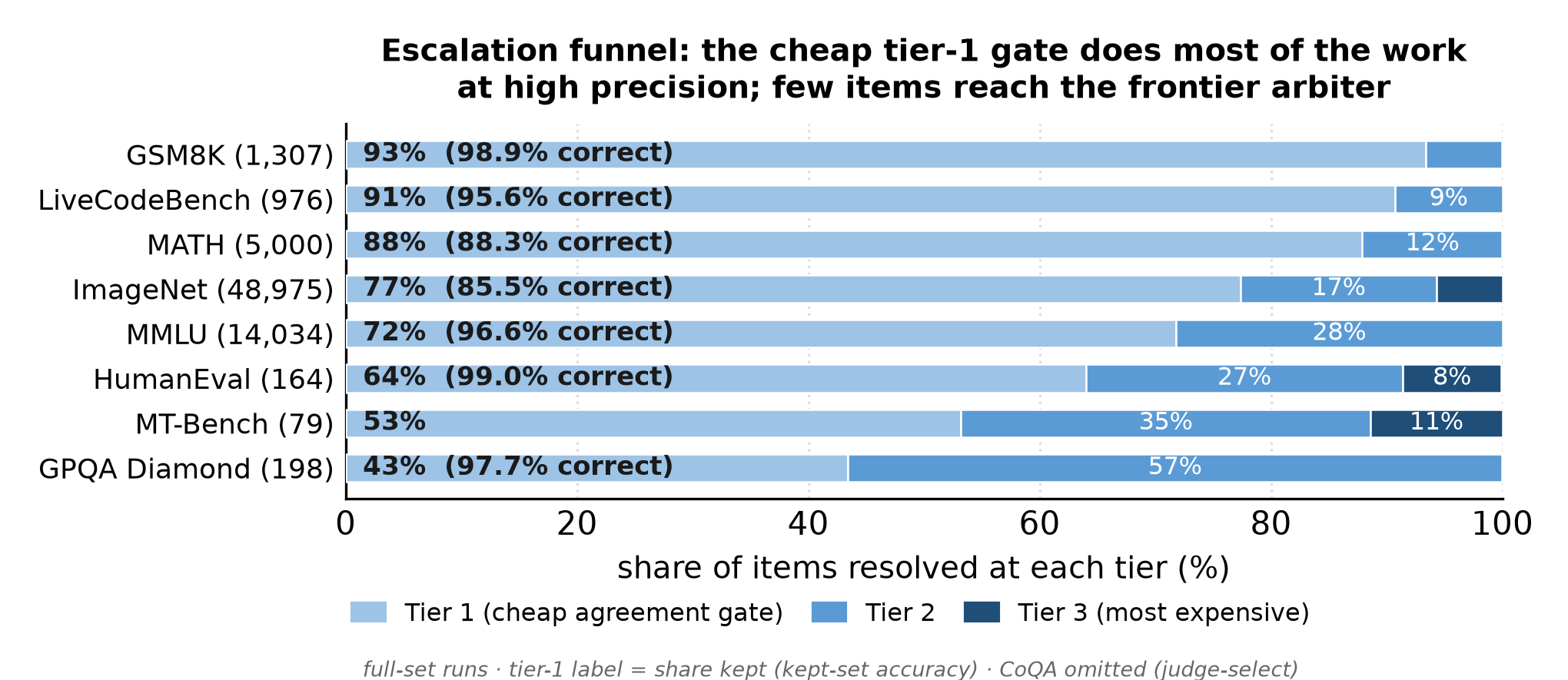}
\caption{Escalation funnel on the eight tiered benchmarks (CoQA uses judge selection): share of items resolved at each tier, with tier-1 kept share and kept-set accuracy annotated. MT-Bench tier-1 quality is a judge score, not an accuracy.}
\label{fig:funnel}
\end{figure}

\begin{figure}[tb]
\centering
\includegraphics[width=\columnwidth]{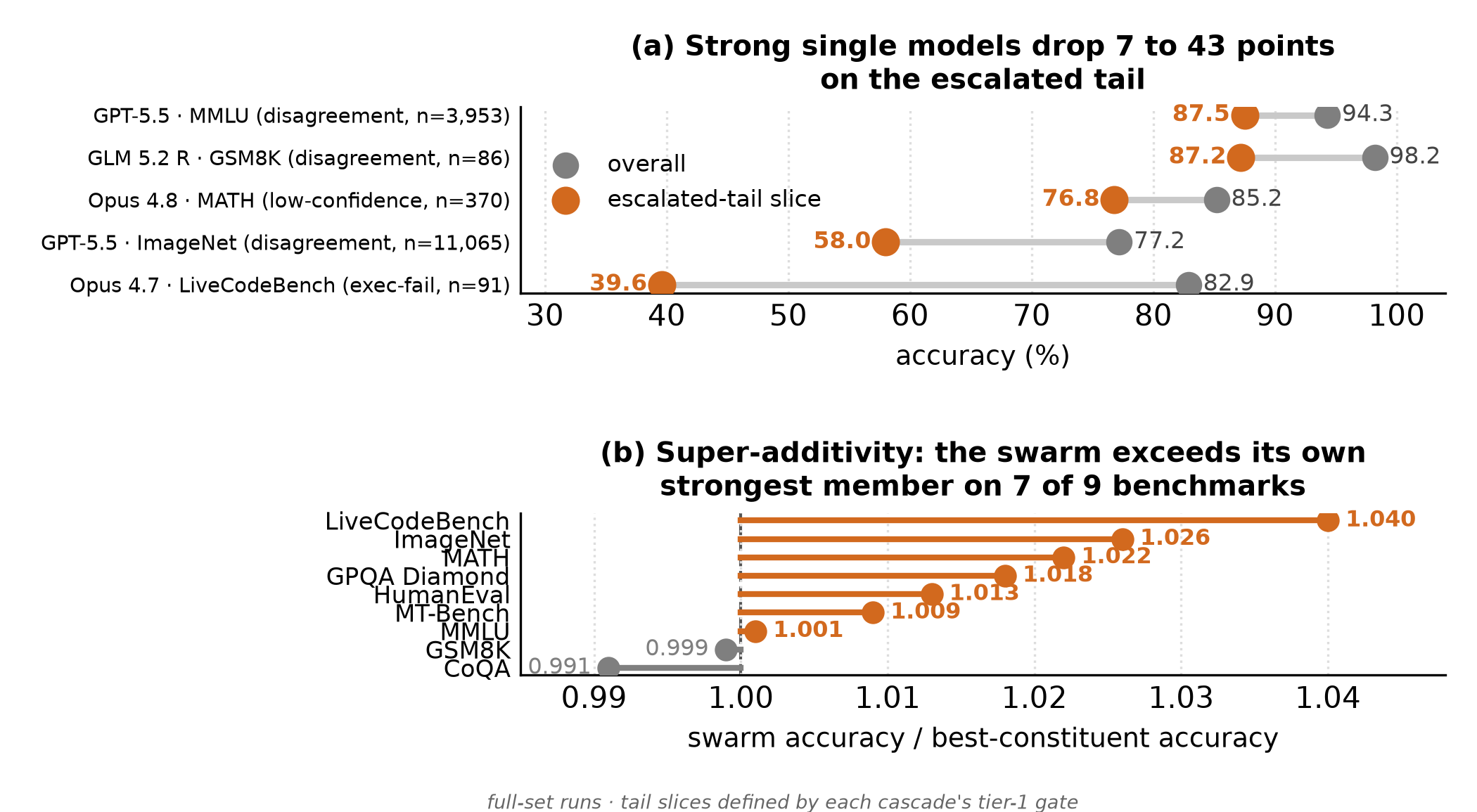}
\caption{(a) Strong single models drop 7 to 43 points on the escalated tail versus overall: gate disagreement isolates genuinely hard items. (b) Swarm/best-constituent accuracy ratio, where values above 1.0 mark super-additive coordination.}
\label{fig:tails}
\end{figure}

Figure~\ref{fig:funnel} shows the funnel: the tier-1 gate resolves 43\% to 93\% of items at 85.5\% to 99.0\% kept-set accuracy, and the most expensive model in each roster sees only 0.7\% to 28\%, the mechanism behind the cost advantage. Escalated items are measurably harder (Figure~\ref{fig:tails}a): even the strongest single models fall well below their overall score there, so producer disagreement is an effective difficulty signal. The same runs show the swarm exceeding its own strongest constituent on seven of nine benchmarks (Figure~\ref{fig:tails}b).

\subsubsection{Official FrugalGPT Comparison.}
\begin{figure}[tb]
\centering
\includegraphics[width=\columnwidth]{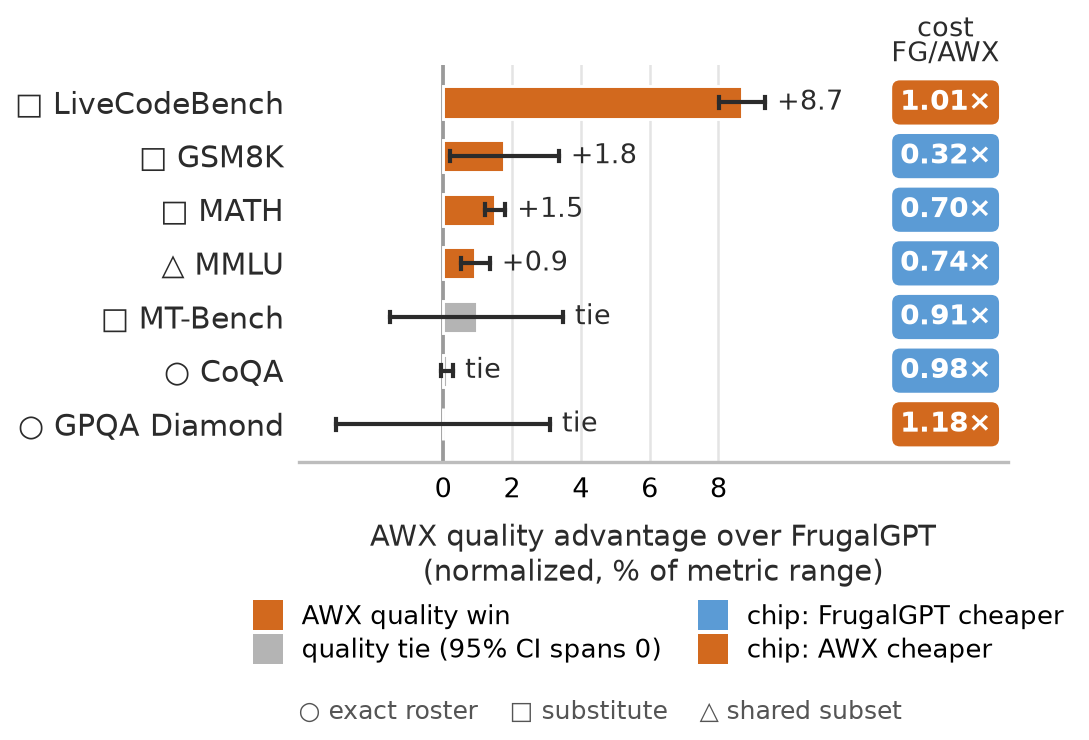}
\caption{Official FrugalGPT vs.\ AWX (five-seed held-out). Bars are AWX's quality advantage over FrugalGPT, normalized per metric with 95\% seed intervals (orange win, gray tie). Each chip is the FrugalGPT/AWX cost ratio (blue FrugalGPT-cheaper, orange AWX-cheaper) and its glyph marks roster status. AWX matches or exceeds FrugalGPT quality on all seven benchmarks and is never beaten at lower cost.}
\label{fig:frugalgpt}
\end{figure}

We separately benchmark the official open-source FrugalGPT implementation~\citep{chen2023frugalgpt} under five group-safe train/validation/test splits, retraining its released DistilBERT scorers and fixing cascade order, thresholds, and budget on validation before the held-out test read. As Figure~\ref{fig:frugalgpt} shows, AWX matches or exceeds FrugalGPT mean quality on all seven benchmarks, including the two exact-roster comparisons (CoQA, GPQA Diamond), and no FrugalGPT policy reaches AWX quality at a lower cost, so it offers cheaper lower-quality operating points but does not Pareto-dominate the training-free gate. The five substitute or shared-subset rows also reflect candidate-pool differences, and HumanEval and ImageNet are excluded (a missing AWX per-problem matrix, and a scorer that cannot inspect image pixels). The full protocol and per-seed results are in the supplementary material.

\subsection{Production Case Studies}
\label{sec:adrel}

\textbf{Ad-relevance.} On the workload of Section~\ref{sec:adrelsetup}, AWX sits on the cost-quality frontier, matching or exceeding every single-shot comparator at roughly half the cost of the strongest, the widest margin in our study. It leads the best single-shot model by about 2.5 points of booked-relevance accuracy and a deployed GPT baseline by about 6.6 points, with Section~\ref{sec:mechanism-ablation} attributing most of the margin to the domain-adapter harness. Absolute booked-relevance scores are withheld as proprietary production data.

\textbf{Image classification.} On the image workload, AWX runs at a 4.3$\times$ cost advantage over single-shot Opus, resolving 89.1\% of images at Tier~1 with only 0.7\% reaching Opus. On disputed scene classifications (28.1\% of images) the swarm is correct 57.6\% against 42.4\% for Opus ($n = 2{,}168$, $p < 10^{-11}$), marketability is statistically level ($p = 0.09$), and disputed scenes carry a 5.8$\times$ cost-accuracy advantage.

\subsection{Swarm Intelligence Metrics}
\label{sec:siq}

The \textbf{Swarm Intelligence Quotient} (SIQ) integrates four emergence indicators:
\begin{equation}
\text{SIQ} = 2\left(\alpha R + \beta J + \gamma(1 - H_n) + \delta S\right)
\label{eq:siq}
\end{equation}
where $R$ is the Kuramoto order parameter, $J$ cluster stability, $H_n$ normalized Shannon entropy, $S$ role specialization, and $(\alpha, \beta, \gamma, \delta) = (0.3, 0.25, 0.25, 0.2)$, and values above 1.0 mark super-additive emergence. Across the nine benchmarks the SIQ ranges from 1.52 to 1.75 (mean 1.69), with accuracy ratios in Figure~\ref{fig:tails}b. Ad-relevance reaches 1.57 (ratio 1.03) and the image domain 1.44.

\subsection{Component Ablation}
\label{sec:mechanism-ablation}

Because each headline cascade is a deterministic composition over cached full-set runs, we ablate it by offline recomposition at full $n$ (65 arms, ad-relevance re-run live), reporting the most informative arm per task in Table~\ref{tab:ablation_combined}.

Value tracks the gate hierarchy (Table~\ref{tab:ablation_combined}): the escalation tail carries the win under weak self-confidence gates (GPQA, MATH), rescues correctly-rejected items under the executable gate (LiveCodeBench), and is purely a cost device under strong consensus (GSM8K). Diversity beats scale wherever testable, as every solo constituent trails its cascade (ImageNet, HumanEval) and the CoQA judge captures only 11\% of oracle-selection headroom. A GSM8K gate-swap arm, a FrugalGPT-style confidence cascade on the same models, loses to AWX consensus (98.16 vs.\ 97.71) at similar cost. For ad-relevance the AWX harness lifts the Nova 2 Lite producer by about 5.5 points with the multi-agent layer adding only $+0.05$ to $+0.15$, so the margin comes chiefly from the domain adapter.

\begin{table}[t]
\centering
\scriptsize
\setlength{\tabcolsep}{4pt}
\begin{tabular}{@{}llrr@{}}
\toprule
\textbf{Benchmark} & \textbf{Component removed} & \textbf{$\Delta$ quality} & \textbf{$\Delta$ cost} \\
\midrule
MMLU          & voter pool $4{\to}2$        & $-0.63^{\ast}$  & $-73\%$ \\
GPQA Diamond  & escalation tail             & $-12.12^{\ast}$ & $-69\%$ \\
GSM8K         & consensus gate (GLM solo)   & $+0.08$         & $+64\%$ \\
MATH          & confidence gate (Kimi solo) & $-1.26^{\ast}$  & $-59\%$ \\
HumanEval     & behavioral gate             & $-1.22$         & $-4\%$  \\
LiveCodeBench & escalation tail             & $-3.69^{\ast}$  & $-96\%$ \\
CoQA (F1)     & judge (best producer only)  & $-0.55^{\ast}$  & $-25\%$ \\
MT-Bench      & GLM mid-tier                & $-0.08$         & $+61\%$ \\
ImageNet      & Opus tier (T3)              & $-0.72^{\ast}$  & $-10\%$ \\
Ad-relevance  & evaluator$+$arbiter         & $-0.05$         & $-89\%$ \\
\bottomrule
\end{tabular}
\caption{Component ablation: the most informative single component removed per task at full $n$. $\Delta$ quality is in the task's metric points, $\Delta$ cost relative to the full cascade. $\ast$ marks paired McNemar (Wilcoxon for CoQA) $p<0.05$, surviving Holm correction.}
\label{tab:ablation_combined}
\end{table}

\subsubsection{PSO-Tuned vs.\ Static Baselines.}
Against the best fixed-roster static cascade, PSO voter selection and gate construction add small but consistent gains (up to $+1.22$pp on the HumanEval behavioral gate and $+0.72$pp on the ImageNet ladder), each measured against a published self-consistency/Mixture-of-Agents baseline~\citep{wang2023selfconsistency,wang2024moa}. Two \emph{trained} routers close the comparison: a FrugalGPT-style scorer (distinct from Figure~\ref{fig:frugalgpt}) never Pareto-dominates the training-free gate, and a MasRouter-style router trails AWX on all nine seed$\times$benchmark rows. A cost-matched self-consistency baseline also trails, by $1.3$pp (GSM8K) and $9.6$pp (GPQA). Per-seed tables for all three baselines are in the supplementary material.

\section{Conclusion}

AWX shows that a training-free swarm of cheap, heterogeneous models, coordinated by stigmergic signals and a cost-aware cascade of task-adapted agreement gates, matches or beats the most accurate single model on all nine public benchmarks (strictly on eight) at a median 34\% of frontier cost, and carries the same advantage into production. One condition explains it: the swarm dominates whenever its cheap gate keeps a subset on which consensus is at least as accurate as the best single model. Gate quality, not spend, is the lever.

\section*{Ethical Statement}
AWX orchestrates existing models at lower cost and energy, with full provenance for auditability. Its main risks, shared-bias reinforcement and low-cost automated decisions at scale, are mitigated by model diversity and the provenance trail.

\bibliography{references}

\appendix
\section*{Appendix}
\addcontentsline{toc}{section}{Appendix}

This appendix contains the full component-ablation grid behind Table~5, threshold-sensitivity and voter-diversity analyses, the production ad-relevance ablation, the measured dominance-condition pairs, the strategy-search protocol, SIQ definitions with per-run components, learned-router and official-FrugalGPT baseline comparisons, and a worked escalation example. All numbers are computed from archived per-item run artifacts released with the code; artifact filenames are given per section.

\section{Full Component-Ablation Grid}
\label{supp:ablation}

The main text's Table~5 reports the single most informative ablation arm per task. Tables~\ref{supp:abl:mmlu}--\ref{supp:abl:imagenet} report \emph{all} arms. Because every headline cascade is a deterministic composition over cached full-set per-item runs, each arm is an offline recomposition at full evaluation-set $n$; the ad-relevance arms (Section~\ref{supp:adrel}) are live re-runs. Each arm is compared pair-wise per item against the FULL cascade on the same item universe with an exact two-sided McNemar test (Wilcoxon signed-rank on per-turn F1 for CoQA and on per-turn judge scores for MT-Bench). Arms marked \emph{partial} are computed only on the subset of escalated items covered by the archived tail run and are not extrapolated; arms that cannot be computed from archives are omitted with the reason recorded in the per-benchmark artifact (\texttt{ablation\_v2/ablation\_\{bench\}.json}). Costs are sums of archived per-item cost fields; on MATH and ImageNet these embed pre-repricing rates, so arm costs are internally comparable within a benchmark but the FULL-arm cost can differ from the canonically re-priced headline.

\paragraph{Multiple comparisons.}
Across the 53 non-reference arms with a defined test, 34 are significant at $p<0.05$ uncorrected and 24 remain significant under Holm--Bonferroni correction over all 53 tests. Every arm starred in Table~5 of the main text survives the correction; the arms that lose significance are the small-$n$ or near-tie arms (e.g., GSM8K voter pairs at $n{=}1{,}308$ with single-item discordance), whose deltas the main text already describes as ties.


\begin{table}[tbp]
\centering\scriptsize
\setlength{\tabcolsep}{3.5pt}
\begin{tabular}{@{}lrrrrr@{}}
\toprule
\textbf{Arm} & \textbf{n} & \textbf{Metric} & \textbf{$\Delta$} & \textbf{$p$} & \textbf{\$/item} \\
\midrule
FULL\_consensus\_gpt\_tail & 14,034 & 94.022 & -- & -- & 0.01469 \\
NO\_GATE\_gpt55\_all & 14,034 & 94.257 & +0.235 & 0.0035 & 0.01157 \\
3v\_gemma4+kimi+llama4m & 14,034 & 93.851 & -0.171 & 0.0071 & 0.01223 \\
3v\_gemma4+kimi+nova\_micro & 14,034 & 93.822 & -0.2 & 0.0003 & 0.00707 \\
3v\_gemma4+llama4m+nova\_micro & 14,034 & 93.68 & -0.342 & 1e-6 & 0.01252 \\
3v\_kimi+llama4m+nova\_micro & 14,034 & 93.844 & -0.178 & 0.0059 & 0.01405 \\
2v\_gemma4+kimi & 14,034 & 93.387 & -0.634 & $<$1e-9 & 0.00404 \\
\bottomrule
\end{tabular}
\caption{MMLU: all ablation arms (paired vs.\ FULL on the common universe). universe = intersection of item ids across all 5 files (gemma4 file has 14,034 rows; others 14,042); 4th voter = Nova Micro raw letter from producer\_answers in the P1 file; its cost is charged as the full P1 best\_of\_n cost\_usd (the archived cascade ran P1 anyway for the confidence pre-flag), so 4-voter-arm costs slightly overstate a pure-Nova-Micro voter.}
\label{supp:abl:mmlu}
\end{table}

\begin{table}[tbp]
\centering\scriptsize
\setlength{\tabcolsep}{4pt}
\begin{tabular}{@{}lrrrrr@{}}
\toprule
\textbf{Arm} & \textbf{n} & \textbf{Metric} & \textbf{$\Delta$} & \textbf{$p$} & \textbf{\$/item} \\
\midrule
FULL & 198 & 87.3737 & -- & 1 & 0.01995 \\
GATE\_0.85 & 198 & 84.3434 & -3.0303 & 0.109 & 0.01379 \\
GATE\_0.75 & 198 & 81.8182 & -5.5556 & 0.0266 & 0.00954 \\
GATE\_0.7 & 198 & 80.8081 & -6.5657 & 0.0106 & 0.00931 \\
GATE\_0.6 & 198 & 79.2929 & -8.0808 & 0.0025 & 0.00820 \\
NO\_GATE\_KIMI\_SOLO & 198 & 75.2525 & -12.1212 & 4e-5 & 0.00620 \\
GPT55\_SOLO & 198 & 85.8586 & -1.5152 & 0.678 & 0.04066 \\
TAIL\_GPT55\_SWAP & 198 & 84.8485 & -2.5253 & 0.383 & 0.04045 \\
\bottomrule
\end{tabular}
\caption{GPQA Diamond: all ablation arms (paired vs.\ FULL on the common universe). Archived gate reconstructed from file: escalate iff kimi\_conf<0.9 (kept confs all $\geq$0.9, escalated all <0.9).; Opus 4.7 tail results exist only for the archived 112 escalated items, so only gates $\leq$0.9 are computable; gates >0.9 are n/a (escalated set would exceed Opus coverage)..}
\label{supp:abl:gpqa}
\end{table}

\begin{table*}[t]
\centering\scriptsize
\setlength{\tabcolsep}{4pt}
\begin{tabular}{@{}lrrrrr@{}}
\toprule
\textbf{Arm} & \textbf{n} & \textbf{Metric} & \textbf{$\Delta$} & \textbf{$p$} & \textbf{\$/item} \\
\midrule
FULL\_3v\_consensus\_glm\_tiebreak & 1,308 & 98.165 & -- & -- & 0.00111 \\
NO\_GATE\_glm\_solo & 1,308 & 98.242 & +0.076 & 1 & 0.00182 \\
2v\_gemma4+llama4\_maverick & 1,308 & 98.089 & -0.076 & 1 & 0.00038 \\
2v\_gemma4+nova\_pro & 1,308 & 98.089 & -0.076 & 1 & 0.00096 \\
2v\_llama4\_maverick+nova\_pro & 1,308 & 97.63 & -0.535 & 0.0156 & 0.00090 \\
GATE\_SWAP\_gemma4\_conf0.99\_glm\_tail & 1,308 & 97.706 & -0.459 & 0.1094 & 0.00014 \\
SOLO\_gemma4 & 1,308 & 97.477 & -0.688 & 0.0225 & 0.00010 \\
SOLO\_llama4\_maverick & 1,308 & 96.636 & -1.529 & 9e-5 & 0.00012 \\
SOLO\_nova\_pro & 1,308 & 93.884 & -4.281 & $<$1e-9 & 0.00055 \\
\bottomrule
\end{tabular}
\caption{GSM8K: all ablation arms (paired vs.\ FULL on the common universe). universe = intersection of item ids across the 3 voter files and the GLM file; answers normalized numerically (strip commas/whitespace) before agreement test.}
\label{supp:abl:gsm8k}
\end{table*}

\begin{table}[tbp]
\centering\scriptsize
\setlength{\tabcolsep}{4pt}
\begin{tabular}{@{}lrrrrr@{}}
\toprule
\textbf{Arm} & \textbf{n} & \textbf{Metric} & \textbf{$\Delta$} & \textbf{$p$} & \textbf{\$/item} \\
\midrule
FULL & 5,000 & 87.02 & -- & 1 & 0.00510 \\
GATE\_0.9 & 5,000 & 86.56 & -0.46 & 0.0165 & 0.00289 \\
GATE\_0.95 & 5,000 & 86.56 & -0.46 & 0.0165 & 0.00289 \\
GATE\_1.0 & 5,000 & 87.02 & -- & 1 & 0.00510 \\
NO\_GATE\_KIMI\_SOLO & 5,000 & 85.76 & -1.26 & 3e-8 & 0.00208 \\
TAIL\_SONNET46 & 5,000 & 86.24 & -0.78 & 0.0002 & 0.00415 \\
TAIL\_GPT55\_PARTIAL & 4,657 & 87.5456 & -0.3221 & 0.0722 & 0.00437 \\
TAIL\_OPUS48\_PARTIAL & 4,762 & 87.3583 & -0.105 & 0.56 & 0.00341 \\
\bottomrule
\end{tabular}
\caption{MATH: all ablation arms (paired vs.\ FULL on the common universe). Item i=1798 (conf=0.5) missing from Opus4.7 tail file: scored wrong with P1-only cost in FULL and gate-sweep arms.; GPT-5.5 (264/607) and Opus 4.8 (369/607) tails are partial-coverage arms: computed on covered subset only, not extrapolated..}
\label{supp:abl:math}
\end{table}

\begin{table}[tbp]
\centering\scriptsize
\setlength{\tabcolsep}{4pt}
\begin{tabular}{@{}lrrrrr@{}}
\toprule
\textbf{Arm} & \textbf{n} & \textbf{Metric} & \textbf{$\Delta$} & \textbf{$p$} & \textbf{\$/item} \\
\midrule
FULL & 164 & 96.34 & -- & -- & 0.00429 \\
NO-BEHAVIORAL-GATE & 164 & 95.12 & -1.22 & 0.5 & 0.00410 \\
NO-CORROBORATION-T2 & 164 & 95.12 & -1.22 & 0.625 & 0.00240 \\
T3-SWAP-OPUS48 & 164 & 95.73 & -0.61 & 1 & 0.00375 \\
SOLO-GPT55 & 164 & 95.73 & -0.61 & 1 & 0.00796 \\
SOLO-DEEPSEEK & 164 & 83.54 & -12.8 & $<$1e-9 & 0.00029 \\
SOLO-KIMI & 164 & 89.63 & -6.71 & 0.001 & 0.00040 \\
\bottomrule
\end{tabular}
\caption{HumanEval: all ablation arms (paired vs.\ FULL on the common universe). p = McNemar exact binomial vs FULL on per-item served answers; margins of 1-2 items are not significant.; NO-BEHAVIORAL-GATE strict variant (keep iff gate\_raw is True, None->escalate): {'metric': 95.12, 'raw': '156/164', 'cost': 0.0041, 'tiers': [110, 40, 14]}.}
\label{supp:abl:humaneval}
\end{table}

\begin{table}[tbp]
\centering\scriptsize
\setlength{\tabcolsep}{4pt}
\begin{tabular}{@{}lrrrrr@{}}
\toprule
\textbf{Arm} & \textbf{n} & \textbf{Metric} & \textbf{$\Delta$} & \textbf{$p$} & \textbf{\$/item} \\
\midrule
FULL & 976 & 90.37 & -- & -- & 0.00768 \\
NO-P2 & 976 & 86.68 & -3.69 & $<$1e-9 & 0.00030 \\
NO-GATE-GEMMA-SOLO & 976 & 86.68 & -3.69 & $<$1e-9 & 0.00031 \\
P2-SWAP-GPT55 & 976 & 89.96 & -0.41 & 0.3877 & 0.00597 \\
SOLO-GPT55 & 976 & 87.19 & -3.18 & 0.0002 & 0.02975 \\
SOLO-OPUS47 & 976 & 87.5 & -2.87 & 0.0008 & 0.02177 \\
\bottomrule
\end{tabular}
\caption{LiveCodeBench: all ablation arms (paired vs.\ FULL on the common universe). Universe = 976 final=OK rows (79 rows ERROR/HTTP-504 excluded, matching archived summary n=976).; FULL mean per-item cost recomputed from the cascade file = its total 7.4942/976 = \$0.00768; the archived summary's cost\_per\_problem field says \$0.00652 (headline) — discrepancy in the archived summary itself; we report the recomputed value..}
\label{supp:abl:lcb}
\end{table}

\begin{table}[tbp]
\centering\scriptsize
\setlength{\tabcolsep}{4pt}
\begin{tabular}{@{}lrrrrr@{}}
\toprule
\textbf{Arm} & \textbf{n} & \textbf{Metric} & \textbf{$\Delta$} & \textbf{$p$} & \textbf{\$/item} \\
\midrule
FULL & 7,983 & 0.8051 & -- & -- & 0.00271 \\
ALWAYS-NOVA & 7,983 & 0.7815 & -0.0236 & $<$1e-9 & 0.00057 \\
ALWAYS-SONNET & 7,983 & 0.7995 & -0.0055 & 0.0003 & 0.00203 \\
ORACLE-SELECT & 7,983 & 0.8509 & +0.0458 & $<$1e-9 & -- \\
\bottomrule
\end{tabular}
\caption{CoQA: all ablation arms (paired vs.\ FULL on the common universe). Producer files ARE present (contrary to prior doubt): Nova Pro per-turn from its cascade P1 checkpoint; Sonnet 4.5 per-turn from Pareto\_plot single-shot checkpoint. Both 7,983 rows, idx 0..7982.; Alignment assumption: select run 'i' and producer 'idx' both enumerate the same 7,983-turn dataset order (same runner); spot-check idx0 predicted 'White' matches across files..}
\label{supp:abl:coqa}
\end{table}

\begin{table}[tbp]
\centering\scriptsize
\setlength{\tabcolsep}{4pt}
\begin{tabular}{@{}lrrrrr@{}}
\toprule
\textbf{Arm} & \textbf{n} & \textbf{Metric} & \textbf{$\Delta$} & \textbf{$p$} & \textbf{\$/item} \\
\midrule
FULL & 158 & 8.937 & -- & -- & 0.00701 \\
SWEEP-T1$\leq$7.5-T2$\leq$7 & 158 & 8.842 & -0.095 & 0.0001 & 0.00490 \\
SWEEP-T1$\leq$8.5-T2$\leq$7 & 158 & 8.937 & -- & 1 & 0.00701 \\
SWEEP-T1$\leq$8-T2$\leq$6.5 & 158 & 8.911 & -0.025 & 0.1026 & 0.00640 \\
SWEEP-T1$\leq$8-T2$\leq$7.5 & 158 & 8.937 & -- & 1 & 0.00701 \\
SKIP-T2 & 158 & 8.854 & -0.082 & 0.0686 & 0.01126 \\
SOLO-GEMMA4 & 158 & 8.608 & -0.329 & $<$1e-9 & 0.00028 \\
SOLO-GLM32K & 158 & 8.861 & -0.076 & 0.1694 & 0.00931 \\
SOLO-OPUS48 & 158 & 8.829 & -0.108 & 0.0618 & 0.02174 \\
\bottomrule
\end{tabular}
\caption{MT-Bench: all ablation arms (paired vs.\ FULL on the common universe). Common question set n=79 (dropped qid(s): [100] — missing from single\_opus48). FULL recomposition reproduces archived summary exactly (8.937, esc\_t2=37, esc\_t3=9, \$0.00701/row).; Escalation rule = min of the two turn judge scores $\leq$ threshold (validated against archived counts)..}
\label{supp:abl:mtbench}
\end{table}

\begin{table}[tbp]
\centering\scriptsize
\setlength{\tabcolsep}{4pt}
\begin{tabular}{@{}lrrrrr@{}}
\toprule
\textbf{Arm} & \textbf{n} & \textbf{Metric} & \textbf{$\Delta$} & \textbf{$p$} & \textbf{\$/item} \\
\midrule
FULL & 48,682 & 79.5571 & -- & 1 & 0.02301 \\
NO\_T3 & 48,682 & 78.8402 & -0.7169 & 8e-38 & 0.02073 \\
NO\_CORROBORATION & 48,682 & 79.5037 & -0.0534 & 0.156 & 0.02301 \\
GPT\_SOLO & 48,682 & 77.2955 & -2.2616 & 2e-87 & 0.03297 \\
OPUS\_SOLO & 48,682 & 77.5461 & -2.011 & 5e-43 & 0.04213 \\
KIMI\_SOLO & 48,682 & 75.3995 & -4.1576 & 1e-217 & 0.01196 \\
GEMMA\_SOLO & 48,682 & 73.52 & -6.0371 & $<$1e-9 & 0.00105 \\
\bottomrule
\end{tabular}
\caption{ImageNet val top-1: all ablation arms (paired vs.\ FULL on the common universe). Universe = 48682 images where all four checkpoints have a usable (non-null/non-ERROR) prediction; duplicates resolved keep-last; ground-truth consistent across all four files (asserted).; Scoring uses the archived per-item correct flag (the original grader, which normalizes labels); exact-string predicted==ground\_truth disagrees with the flag on 83 of 194728 model-item pairs..}
\label{supp:abl:imagenet}
\end{table}

\section{Threshold Sensitivity}
\label{supp:threshold}

The gate thresholds selected by the strategy search are locally flat or monotone, not knife-edge. On MATH (Table~\ref{supp:abl:math}), sweeping the Kimi confidence gate from $0.99$ to $1.0$ leaves accuracy unchanged (87.02\%) at $+\$0.000002$/item, while relaxing to $0.90$--$0.95$ loses $0.46$ points for $-43\%$ cost: the operating point sits on a plateau whose only cliff is toward under-escalation. On GPQA (Table~\ref{supp:abl:gpqa}), tightening the gate from the deployed $0.9$ cutoff monotonically trades accuracy for cost ($-3.0$ points at $0.85$ down to $-8.1$ at $0.6$), with no non-monotonic surprises. On MT-Bench (Table~\ref{supp:abl:mtbench}), loosening either judge threshold by $0.5$ changes nothing (integer-valued judge scores), and tightening tier~1 to $\leq 7.5$ costs $-0.095$ ($p{=}10^{-4}$): the deployed thresholds are the cheapest setting on their plateau.

\section{Voter Diversity}
\label{supp:voters}

Removing any single voter from the MMLU 4-voter consensus costs $0.17$--$0.34$ points (all McNemar $p<0.01$; Table~\ref{supp:abl:mmlu}), and collapsing to the best 2-voter pair costs $0.63$ points; the cheapest 3-voter arm (Gemma~4 + Kimi + Nova Micro) retains $93.8\%$ at $37\%$ lower cost and is the value point. On GSM8K, two of the three 2-voter pairs tie the full trio within noise, but the pair excluding Gemma~4 loses $0.54$ points ($p{=}0.016$): which voters, not how many, carries the signal. Every solo constituent is significantly worse than its cascade on every benchmark with headroom, consistent with the main text's producer-diversity claim.

\section{Production Ad-Relevance Ablation}
\label{supp:adrel}

The production cascade (Nova 2 Lite producer, Gemma 3 27B evaluator, Claude Opus 4.7 arbiter) was re-run live on the 1{,}000-search golden set under four configurations, with per-producer labels and arbiter decisions persisted (Table~\ref{supp:tab:adrel}). The full cascade is the only configuration that judges every booked ad; each reduced arm drops one or two to unjudged, which the primary metric BRS\_Total penalizes by design. A separate control run of Nova 2 Lite through the same single-shot harness as all baseline models isolates the pipeline's contribution: the AWX domain-adapter pipeline alone lifts the producer by about $+5.5$ points, and the multi-agent layer adds the final $+0.05$ to $+0.15$ plus full judged coverage. The producer's labels agree with the final output on 99.9\% of ads, and the Opus arbiter fires so rarely that removing it does not measurably change cost. Absolute booked-relevance scores and per-search costs are withheld as proprietary production data.

\begin{table}[t]
\centering\small
\begin{tabular}{@{}lr@{}}
\toprule
\textbf{Configuration} & \textbf{$\Delta$ quality vs full} \\
\midrule
Full production cascade & reference \\
$-$ evaluator (Gemma 3 27B) & $-0.15$ \\
$-$ arbiter (Opus 4.7) & $-0.05$ \\
Nova 2 Lite solo, in-pipeline & $-0.05$ \\
Nova 2 Lite raw single-shot & $-5.5$ \\
\bottomrule
\end{tabular}
\caption{Ad-relevance ablation on the 1{,}000-search golden set: quality change relative to the full production cascade (BRS\_Total). The full cascade is the only arm with all booked ads judged. Absolute scores and costs are withheld as proprietary production data. Artifacts: \texttt{ablation\_adrelevance.csv}, \texttt{ablation\_adrelevance\_attribution.csv}, \texttt{nova2lite\_metrics.json}.}
\label{supp:tab:adrel}
\end{table}

\section{Dominance-Condition Data}
\label{supp:dominance}

Table~\ref{supp:tab:dom} lists the measured pairs behind the dominance condition of the main text (Section~5.1 and Table~4): the tier-1 consensus accuracy and the best single model's accuracy restricted to the same kept subset, joined per item on archived full-set runs.

\begin{table}[t]
\centering\small
\setlength{\tabcolsep}{4pt}
\begin{tabular}{@{}lrrrr@{}}
\toprule
\textbf{Benchmark} & \textbf{$n_{\text{kept}}$} & \textbf{Consensus} & \textbf{Best single} & \textbf{Join} \\
\midrule
ImageNet      & 37,861 & 85.60 & 84.17 & 100\% \\
LiveCodeBench &    885 & 95.59 & 94.35 & 100\% \\
HumanEval     &    105 & 99.05 & 99.05 & 100\% \\
GSM8K         &  1,221 & 98.94 & 99.02 & 99.5\% \\
MMLU          & 10,081 & 96.58 & 96.91 & 100\% \\
GPQA Diamond  &     86 & 97.67 & 100.00 & 100\% \\
MT-Bench      &     42 & 9.298 & 9.286 & 100\% \\
\bottomrule
\end{tabular}
\caption{Kept-subset pairs (best single per Table~3 of the main text; MT-Bench values are judge scores). MATH and CoQA are excluded: no archived full-set per-item run exists for their best single models (Opus 4.8 on MATH was run on a 499-item sample; CoQA's select-run artifacts store only final answers).}
\label{supp:tab:dom}
\end{table}

\section{Legacy Roster-Independent Mechanism Ablation}
\label{supp:legacy}

An earlier leave-one-out study ablated the generic AWX engine (three diverse open-weight producers, evaluator, arbiter: Llama 3.3 70B + Nova Pro + Gemma 3 27B $\to$ Kimi K2.5 $\to$ DeepSeek R1; vision roster Nova Pro + Gemma 3 27B + Pixtral Large $\to$ Qwen3-VL) on 300 items per benchmark at temperature 0.0 (Table~\ref{supp:tab:legacy}). It covers per-item mechanisms that the headline component ablation cannot cleanly isolate, and its conclusions agree: producer diversity dominates quality, the consensus vote chiefly controls cost (except on vision, where it also carries quality), multi-round escalation matters on the harder vision task, and forwarded context is a measured null.

\begin{table}[t]
\centering\footnotesize
\setlength{\tabcolsep}{2.6pt}
\begin{tabular}{@{}lrrrr@{}}
\toprule
\textbf{Mechanism removed} & \textbf{GSM8K} & \textbf{MMLU} & \textbf{CoQA} & \textbf{ImageNet} \\
\midrule
Full engine (absolute) & 97.3 & 88.3 & 71.8 & 66.0 \\
\midrule
$-$~Producer diversity & $-0.33$ & $-5.66^{\ast}$ & $-4.05^{\ast}$ & $-20.67^{\ast}$ \\
$-$~Consensus vote & $-0.33$ & $+0.00$ & $-0.98$ & $-12.33^{\ast}$ \\
$-$~Multi-round escalation & $-0.33$ & $-2.33$ & $-0.55$ & $-6.67^{\ast}$ \\
$-$~Forwarded context & $+0.67$ & $-0.33$ & $+0.16$ & $-2.00$ \\
\bottomrule
\end{tabular}
\caption{Legacy leave-one-out engine ablation (300 items each, all-open rosters, exact McNemar $^{\ast}p<0.05$). Absolute scores differ from the headline cascades because the roster and gates differ; the study measures relative mechanism effects.}
\label{supp:tab:legacy}
\end{table}

\section{Weaving Stopping-Criterion Study}
\label{supp:weaving}

The deficit-guided stopping criterion of progressive signal weaving was validated on an event-discovery domain with gap-aware re-production. Allowing one extra weaving round beyond the Lyapunov stopping point adds little to saturated content (New York grows $4.5\%$ as $V$ nears saturation) but sharply expands under-explored content (Los Angeles $+16.8\%$, East Rutherford $+102\%$, Mexico City $+278\%$): the criterion withholds computation where content is saturated while permitting exploration where deficits remain, which is precisely its design intent.

\section{Strategy-Search Protocol}
\label{supp:pso}

The \texttt{EvolutionarySelector} searches an eight-dimensional configuration space with particle swarm optimization: strategy identifier $[0,5]$ (Progressive, Consensus, Polyphonic, Debate, Cascade, Hybrid), model temperature $[0,1.5]$, blend factor $[0,1]$, deficit threshold $[0,1]$, diversity weight $[0,1]$, cost weight $[0,1]$, agent count $[1,50]$, and maximum rounds $[1,10]$. It uses 20 particles with Clerc--Kennedy constriction ($c_1{=}c_2{=}2.05$, $\phi{=}4.1$), a ring topology in which each particle takes its local best from itself and two neighbors, and two heuristic cold-start rounds before the swarm has signal. Fitness is $f = q\cdot r / \max(c, 0.001)$ with $q$ quality, $r$ call reliability, and $c$ cost in USD. For the public benchmarks the search operates over \emph{cached} full-set producer runs, so candidate gates and thresholds are evaluated by recomposition without new model calls; gates never observe ground truth, and the selected configuration is then confirmed by a single deterministic full-set run reported in the main text.

\section{Dynamic Agent-Count Estimation}
\label{supp:agentcount}

Agent counts are estimated per task as $\mathrm{clamp}_{[4,500]}\!\big((b_d + \ell + u + k + e)\cdot s\big)$, where $b_d$ is a per-domain base count, $\ell$ a task-text length bonus, $u$ a quantity-signal bonus (0--6, from explicit counts in the request), $k$ a complexity-keyword score (positive for terms such as \emph{compare}, \emph{comprehensive}; negative for \emph{simple}, \emph{brief}), $e$ a multi-entity bonus (e.g., \emph{rank}, \emph{across}), and $s$ a per-strategy multiplier. The estimator returns the contributing factors alongside the count for auditability.

\section{SIQ Definitions and Per-Run Components}
\label{supp:siq}

$\text{SIQ} = 2(\alpha R + \beta J + \gamma(1-H_n) + \delta S)$ with $(\alpha,\beta,\gamma,\delta) = (0.3, 0.25, 0.25, 0.2)$. For benchmark cascades, $R$ is the Kuramoto order parameter $|\,\mathrm{mean}(e^{2\pi i q})\,|$ over the member solo-quality vector $q$ (answer-serving roster members, each scored on the full set), $H_n$ the normalized Shannon entropy of $q$ over five equal quality bins, $J$ cluster stability, and $S$ role specialization; the accuracy ratio is cascade quality over $\max(q)$. The production ad-relevance deployment computes the same functional from live internal signals instead (per-judgment confidence synchrony for $R$, label-distribution entropy for $H_n$), which is why its components are not comparable cell-for-cell with the benchmark rows in Table~\ref{supp:tab:siq}.

\begin{table}[t]
\centering\footnotesize
\setlength{\tabcolsep}{4pt}
\begin{tabular}{@{}lrrrrr@{}}
\toprule
\textbf{Run} & \textbf{SIQ} & \textbf{$R$} & \textbf{$1{-}H_n$} & \textbf{$J$} & \textbf{Ratio} \\
\midrule
MT-Bench      & 1.750 & 0.998 & 1.000 & 0.5 & 1.009 \\
GSM8K         & 1.747 & 0.995 & 1.000 & 0.5 & 0.999 \\
ImageNet      & 1.746 & 0.994 & 1.000 & 0.5 & 1.026 \\
LiveCodeBench & 1.746 & 0.993 & 1.000 & 0.5 & 1.040 \\
GPQA Diamond  & 1.745 & 0.991 & 1.000 & 0.5 & 1.018 \\
HumanEval     & 1.731 & 0.969 & 1.000 & 0.5 & 1.013 \\
CoQA          & 1.630 & 0.994 & 0.605 & 0.667 & 0.991 \\
Ad-relevance  & 1.573 & 0.949 & 0.208 & 0.999 & 1.029 \\
MMLU          & 1.565 & 0.924 & 0.720 & 0.5 & 1.000 \\
MATH          & 1.518 & -- & -- & -- & 1.022 \\
\bottomrule
\end{tabular}
\caption{SIQ components for all ten runs ($S{=}1.0$ throughout; ratio = swarm/best-constituent quality). MATH's component split was not archived; its composite SIQ is recorded with the run. Ad-relevance components use the live internal-signal basis; its ratio uses the best archived member solo (Gemma 3 27B).}
\label{supp:tab:siq}
\end{table}

\section{Cost Sensitivity}
\label{supp:costsens}

The Pareto verdicts are robust to price changes. On seven of nine benchmarks the strongest frontier comparator costs $1.7\times$ to $2.9\times$ the AWX cascade per item (GPQA $2.05\times$, GSM8K $2.0\times$, MATH $2.03\times$, HumanEval $1.86\times$, CoQA $1.70\times$, MT-Bench $2.64\times$, ImageNet $2.90\times$), so the both-axes result survives any adverse repricing smaller than those factors. The two thin cases are stated in the main text: MMLU's cost edge over GPT-5.5 is $1.03\times$ (an accuracy tie with a marginal cost edge), and on LiveCodeBench the cheaper comparator (Gemma 4) is $3.5$ points less accurate while the accuracy-competitive comparators cost $4$--$6\times$ more.

\section{Learned-Router and FrugalGPT Baselines}
\label{supp:router}

We train a FrugalGPT-style router, a logistic scorer over cheap-tier features (the vote-agreement pattern of top-vote count, distinct answers, and unanimity bit, plus producer confidence, question length, and subject), on the dev half of each 50/50 split to predict cheap-arm correctness, escalating when predicted $P(\text{correct}) < \tau$, with $\tau$ chosen on the dev half at two operating points (cost-matched and accuracy-matched to the gate arm) and evaluated on the held-out test half, with three seeds (7, 13, 42). Producers and the strong escalation arm are identical to the gate arm in every comparison, so the \emph{only} difference is the escalation decision. Table~\ref{supp:tab:router} reports every seed. The trained router never \emph{Pareto-dominates} the training-free agreement gate: at matched cost it never exceeds gate accuracy on any seed of any benchmark. On MMLU and GSM8K it is within $\pm 0.05$pp at near-identical cost (the router re-discovers unanimity: its top coefficient is the unanimity bit), and on MATH the cost-matched router trails the gate on all three seeds. Accuracy exceedances appear only in the accuracy-matched arms (MATH seed 42, $+0.44$pp at $2.2\times$ the gate's cost, and MMLU seed 42, $+0.02$pp), never at matched cost. The routable signal is the agreement pattern itself, and learning adds nothing at matched cost once the gate observes it. Costs in this section use the canonical repricing basis throughout (Llama 4 Maverick at \$0.00048/item), matching the main text rather than the archived-log basis of the \S1 ablation tables.

\paragraph{Model-selection router (MasRouter-style).} We additionally compare against the LLM-selection core of MasRouter (Yue et al.\ 2025, arXiv:2502.11133) in its stronger supervised form: per-model correctness predictors $p_m(x)$ (logistic regression on MiniLM query embeddings, trained on the dev half over the full single-shot pool: the per-item-archived subset of the comparator pool, with 5 models on MMLU, 7 on GSM8K, and 12 on MATH, in a reimplementation of the selection layer rather than MasRouter's released weights) route each query to $\arg\max_m [p_m(x) - \lambda\,c_m]$, with $\lambda$ swept on the dev half and fixed at cost-matched and accuracy-matched operating points before touching the test half (Table~\ref{supp:tab:masrouter}). The router selects \emph{one} model per query, while AWX runs its deployed cascade. On all nine seed$\times$benchmark combinations AWX's held-out accuracy exceeds the router's at the cost-matched point (MMLU $+0.26$ to $+1.00$pp, GSM8K $+0.15$ to $+0.46$pp, MATH $+0.33$ to $+1.07$pp), and the router's accuracy-matched arm loses $0.5$--$2.7$pp of its dev accuracy on the test half on every seed. The structural reason is visible in the ceilings: a single-model router cannot exceed its best pool member (GPT-5.5 at 94.26\% on MMLU, GLM~5.2R at 98.24\% on GSM8K), whereas consensus-plus-escalation can exceed it (MMLU held-out halves and GSM8K seed 42) and elsewhere matches the ceiling at a fraction of the ceiling model's cost. The router does find cheaper operating points at lower accuracy, trading on a different part of the frontier, but at AWX's cost budget it never reaches AWX's accuracy. This covers MasRouter's model-selection layer, not its full multi-agent-construction pipeline (collaboration-mode selection, VAE task classifier), which requires live multi-agent rollouts we cannot recompose from archives. Script: \texttt{benchmarks/masrouter\_baseline.py} with \texttt{all-MiniLM-L6-v2} embeddings.

\begin{table}[t]
\centering\scriptsize
\setlength{\tabcolsep}{3pt}
\begin{tabular}{@{}llrrr@{}}
\toprule
\textbf{Bench} & \textbf{Seed} & \textbf{AWX test} & \textbf{Router$_{\text{cost}}$} & \textbf{Router$_{\text{acc}}$} \\
\midrule
MMLU  & 7  & \textbf{94.43} @ .01100 & 93.43 @ .00773 & 91.76 @ .00343 \\
MMLU  & 13 & \textbf{94.27} @ .01118 & 94.01 @ .00996 & 92.36 @ .00355 \\
MMLU  & 42 & \textbf{94.34} @ .01110 & 93.72 @ .00889 & 91.76 @ .00357 \\
GSM8K & 7  & \textbf{98.16} @ .00115 & 98.01 @ .00096 & 97.70 @ .00008 \\
GSM8K & 13 & \textbf{98.16} @ .00121 & 97.70 @ .00108 & 97.24 @ .00008 \\
GSM8K & 42 & \textbf{98.62} @ .00100 & 98.16 @ .00014 & 98.01 @ .00008 \\
MATH  & 7  & \textbf{86.93} @ .00474 & 86.48 @ .00436 & 84.06 @ .00037 \\
MATH  & 13 & \textbf{86.73} @ .00492 & 86.40 @ .00417 & 84.60 @ .00046 \\
MATH  & 42 & \textbf{87.51} @ .00466 & 86.44 @ .00415 & 84.02 @ .00041 \\
\bottomrule
\end{tabular}
\caption{MasRouter-style learned model-selection router (supervised LLM-selection core, MiniLM embeddings, full single-shot pool) vs.\ the deployed AWX cascade on held-out test halves, reported as accuracy \% @ \$/item. Router$_{\text{cost/acc}}$ = dev-selected cost-matched / accuracy-matched $\lambda$. AWX's held-out accuracy is highest on all nine rows.}
\label{supp:tab:masrouter}
\end{table}

\begin{table}[t]
\centering\scriptsize
\setlength{\tabcolsep}{3pt}
\begin{tabular}{@{}llrrrr@{}}
\toprule
\textbf{Bench} & \textbf{Seed} & \textbf{Gate test} & \textbf{\$/it} & \textbf{Router$_{\text{cost}}$} & \textbf{Router$_{\text{acc}}$} \\
\midrule
MMLU  & 7  & 94.23 & .00741 & 94.23 @ .00740 & 94.23 @ .00740 \\
MMLU  & 13 & 94.00 & .00743 & 93.97 @ .00739 & 93.97 @ .00739 \\
MMLU  & 42 & 94.04 & .00738 & 93.99 @ .00720 & 94.06 @ .00746 \\
GSM8K & 7  & 98.01 & .00106 & 98.01 @ .00103 & 98.01 @ .00103 \\
GSM8K & 13 & 97.55 & .00119 & 97.55 @ .00119 & 97.55 @ .00119 \\
GSM8K & 42 & 98.32 & .00102 & 98.32 @ .00100 & 98.32 @ .00100 \\
MATH  & 7  & 87.08 & .00546 & 86.80 @ .00515 & 86.80 @ .00569 \\
MATH  & 13 & 87.16 & .00553 & 87.12 @ .00424 & 87.08 @ .00960 \\
MATH  & 42 & 87.40 & .00559 & 87.12 @ .00524 & 87.84 @ .01238 \\
\bottomrule
\end{tabular}
\caption{Training-free agreement gate vs.\ trained logistic router on held-out test halves (same producers and same strong arm, with MMLU gate = 4-voter unanimity $\to$ GPT-5.5, GSM8K = 3-voter unanimity $\to$ GLM 5.2R, and MATH = Kimi conf $\geq 0.99 \to$ Opus 4.6). Router$_{\text{cost/acc}}$ = accuracy @ \$/item at the dev-selected cost-matched / accuracy-matched threshold. Script: \texttt{benchmarks/learned\_router\_baseline.py}.}
\label{supp:tab:router}
\end{table}

\paragraph{Official open-source FrugalGPT.}
The feature-based control above is not the released FrugalGPT system. We therefore also run the official code at commit \texttt{2b23e6a}, with its per-model DistilBERT answer scorers and cascade optimizer, under five group-safe 50/25/25 splits (seeds 7, 17, 29, 43, 61). Scorers fit on training only; model order, thresholds, and budget feasibility use validation only; test labels report the fixed policy. Table~\ref{supp:tab:officialfrugal} gives the operating point nearest the AWX validation budget that is feasible on all five seeds. The quality interval is a Student-$t$ interval over seed-level differences; every seed artifact also stores a paired cluster bootstrap. MT-Bench cost is per two-turn conversation and includes AWX's online routing-judge calls. Exact-roster and qualified sensitivity rows are separated explicitly, and no HumanEval or ImageNet point is imputed.

\begin{table*}[t]
\centering
\small
\setlength{\tabcolsep}{4.2pt}
\begin{tabular}{lcccccc}
\toprule
Dataset & Roster & Budget & FrugalGPT & AWX & $\Delta$ quality [95\% seed CI] & Cost ratio \\
\midrule
CoQA          & E & $1.00\times$ & 0.7997 & 0.8008 & $-0.0011\;[-0.0029,\;0.0006]$ & $0.982\times$ \\
GPQA Diamond  & E & $1.25\times$ & 86.9388 & 86.9388 & $\phantom{-}0.0000\;[-3.1030,\;3.1030]$ & $1.184\times$ \\
GSM8K         & S & $1.00\times$ & 96.3190 & 98.0982 & $-1.7791\;[-3.3585,\;-0.1998]$ & $0.318\times$ \\
LiveCodeBench & S & $1.00\times$ & 81.6393 & 90.3279 & $-8.6885\;[-9.3519,\;-8.0251]$ & $1.008\times$ \\
MATH          & S & $1.00\times$ & 85.6090 & 87.1154 & $-1.5064\;[-1.7998,\;-1.2130]$ & $0.703\times$ \\
MMLU          & U & $1.00\times$ & 93.6791 & 94.6132 & $-0.9341\;[-1.3532,\;-0.5150]$ & $0.743\times$ \\
MT-Bench      & S & $1.00\times$ & 8.7375 & 8.8250 & $-0.0875\;[-0.3133,\;0.1383]$ & $0.914\times$ \\
\bottomrule
\end{tabular}
\caption{Official FrugalGPT vs.\ AWX on five repeated train/validation/test splits. E = exact roster, S = documented substitute roster, and U = shared-model subset. The operating point is the validation-nearest budget feasible on all seeds; the cost ratio is the mean of paired seed-level FrugalGPT/AWX ratios. Quality units follow each benchmark (CoQA F1 in $[0,1]$, MT-Bench in $[0,10]$, all others percentage points).}
\label{supp:tab:officialfrugal}
\end{table*}

For the requested matched-cost view, the best mean quality differences among validation-fixed policies at or below AWX cost are CoQA $-0.0011$ at $0.982\times$, GSM8K $-1.1656$ at $0.174\times$, LiveCodeBench $-9.4262$ at $0.837\times$, MATH $-1.2821$ at $0.939\times$, MMLU $-0.9054$ at $0.606\times$, and MT-Bench $-0.0875$ at $0.914\times$; GPQA has no all-seed feasible matched-cost point. For the matched-quality view, GPQA alone reaches AWX mean quality, at $1.184\times$ cost (an 18.4\% increase); no dataset yields an observed cost reduction at or above AWX mean quality. These are descriptive post-hoc summaries over policies already fixed on validation, never threshold choices made from test labels.

\paragraph{Cost-matched self-consistency.} A same-model sampling baseline answers whether AWX's gains come merely from spending the budget on more votes. Two structural constraints scope it: on GSM8K the best single (GLM 5.2R at \$0.0018/q) costs twice AWX's entire per-item budget for one call, and GPT-5.5's reasoning endpoint rejects temperature, so it cannot be sampled. The cost-matched arms therefore use the strongest affordable singles. On GSM8K, Llama 4 Maverick with $k{=}7$ votes at temperature $0.7$ (majority over normalized answers at \$0.00083/q against AWX's \$0.0009) scores $96.89\%$: $+0.25$pp over its own single shot and $1.27$pp below AWX at the same budget. On GPQA, Kimi K2.5 (the AWX tier-1 producer) with $k{=}3$ votes at temperature $0.7$ (${\approx}\$0.019$/q, three times its measured single-call cost, against AWX's \$0.0199) scores $77.78\%$: $+2.5$pp over its single shot and $9.6$pp below AWX. Same-model sampling recovers a small fraction of the cascade's margin at matched cost. The diversity that carries the gate signal lives across models rather than across samples of one model, consistent with the voter-diversity ablation of \S3. Artifacts: \texttt{sc\_gsm8k\_llama4m\_k7.jsonl} and \texttt{sc\_gpqa\_kimi\_k3.jsonl} in \texttt{selfconsistency/}, with the script \texttt{selfconsistency\_baseline.py}.

\section{Worked Escalation Example}
\label{supp:example}

GSM8K item 409 illustrates why the unanimity gate beats majority vote. Gemma 4 and Nova Pro both answered 42{,}000 (identically wrong), while Llama 4 Maverick answered 168{,}000 (correct): a majority vote would have committed the correlated error, but the all-agree gate detected non-unanimity, escalated, and the GLM 5.2 reasoning tiebreak returned the correct answer. Correlated wrong answers among cheap producers are exactly the failure mode that unanimity-plus-strong-tiebreak absorbs and majority voting cannot.

\section{Artifact Index}
\label{supp:artifacts}

Released with the code (paths relative to the code supplement root): \texttt{ablation\_v2/} (nine per-benchmark ablation JSONs with sources and caveats, recomposition scripts, combined CSV, ad-relevance CSVs, Nova 2 Lite control metrics); per-benchmark run summaries with SIQ blocks under \texttt{results\_summaries/}; curated per-item checkpoints for the headline cascades and their single-shot comparators under \texttt{percheckpoints/}; and the canonical rate card at \texttt{repricing/\_shared/cost\_map\_models.csv}. Every headline number is reproduced offline by \texttt{recompute\_scores.py} (Table 3), \texttt{recompute\_gate\_hierarchy.py} (Table 4 and the dominance pairs), and \texttt{recompute\_ablation.py} (Table 5). The official-FrugalGPT package is under \texttt{baselines/frugalgpt\_publication/}; it contains exact split manifests, the declared model roster and cost basis, the pinned upstream commit and compatibility patch, seed-level metrics and confidence intervals, and SHA-256 checksums, reproduced by \texttt{baselines/frugalgpt\_report.py}. Self-consistency caches are under \texttt{selfconsistency/} and the learned-router and MasRouter-style scripts under \texttt{benchmarks/}.

\end{document}